\documentclass[letterpaper, 10 pt, journal, twoside]{IEEEtran}   
\usepackage{amsmath}
\usepackage{bm}

\IEEEoverridecommandlockouts                              

\DeclareMathOperator*{\argmax}{arg\,max}
\DeclareMathOperator*{\argmin}{arg\,min}

\newcommand*{\V}[1]{\bm{#1}}

\usepackage{graphicx} 
\usepackage{amsmath} 
\usepackage{amssymb}  
\usepackage{mathtools}  
\usepackage{siunitx}  

\usepackage{algorithm}
\usepackage{algorithmic}
\usepackage{url}
\usepackage{multirow,bigdelim}
\usepackage[table]{xcolor}

\title{ConFusion: Sensor Fusion for Complex Robotic Systems using Nonlinear Optimization}

\author{Timothy Sandy, Lukas Stadelmann, Simon Kerscher and Jonas Buchli
\thanks{Manuscript received: Sept, 10, 2018; Revised Nov, 6, 2018; Accepted Jan, 2, 2019.}
\thanks{This paper was recommended for publication by Editor Eric Marchand upon evaluation of the Associate Editor and Reviewers' comments. 
This research was supported by the Swiss National Science Foundation through the National Centre of Competence in Research Digital Fabrication (\#51NF40\_141853) and a Professorship Award to Jonas Buchli (\#PP00P2\_138920).} 
\thanks{The authors are with the Agile \& Dexterous Robotics Lab, ETH Zurich, Switzerland. {\tt\small tsandy@ethz.ch}}%
\thanks{Digital Object Identifier (DOI): see top of this page.}
}

\begin{document}

\maketitle

\begin{abstract}
We present ConFusion, an open-source package for online sensor fusion for robotic applications. ConFusion is a modular framework for fusing measurements from many heterogeneous sensors within a moving horizon estimator. ConFusion offers greater flexibility in sensor fusion problem design than filtering-based systems and the ability to scale the online estimate quality with the available computing power. We demonstrate its performance in comparison to an iterated extended Kalman filter in visual-inertial tracking, and show its versatility through whole-body sensor fusion on a mobile manipulator.
\end{abstract}

\section{Introduction}
\IEEEPARstart{S}{ensor} fusion is a valuable tool in the roboticist's toolbox. As the complexity and state dimensionality of robots increase, information from numerous sensors must be considered to determine the full state of the robot for use in motion control. While general methods for sensor fusion are well-established, the design and implementation of state estimators for complex robots is tedious, making it difficult to easily add and remove sensors or to investigate alternative state representations for different tasks. In this work, we look to develop a framework for online sensor fusion that supports a wide range of robots and sensors and provides the flexibility and modularity necessary to easily build and evaluate state estimators for complex robots.

This paper introduces ConFusion, an open-source C++ package for online sensor fusion. ConFusion implements a moving horizon estimator (MHE) to optimize over a sliding batch of states to generate high-accuracy robot state estimates for use in real-time control systems. The effort required to implement a sensor fusion problem is no more than would be required for an extended Kalman filter, but ConFusion's batch-based approach and use of nonlinear optimization provides more flexibility in terms of state estimator design and the ability leverage additional computing resources to improve estimate quality by optimizing over a larger batch of states. Our modular framework allows for the easy incorporation of new sensors and estimated parameters and the ability to easily run offline batch calibration problems using the same state and measurement models.

\begin{figure}
	\centering
	\includegraphics[trim=150 300 500 100,clip,width=\columnwidth]{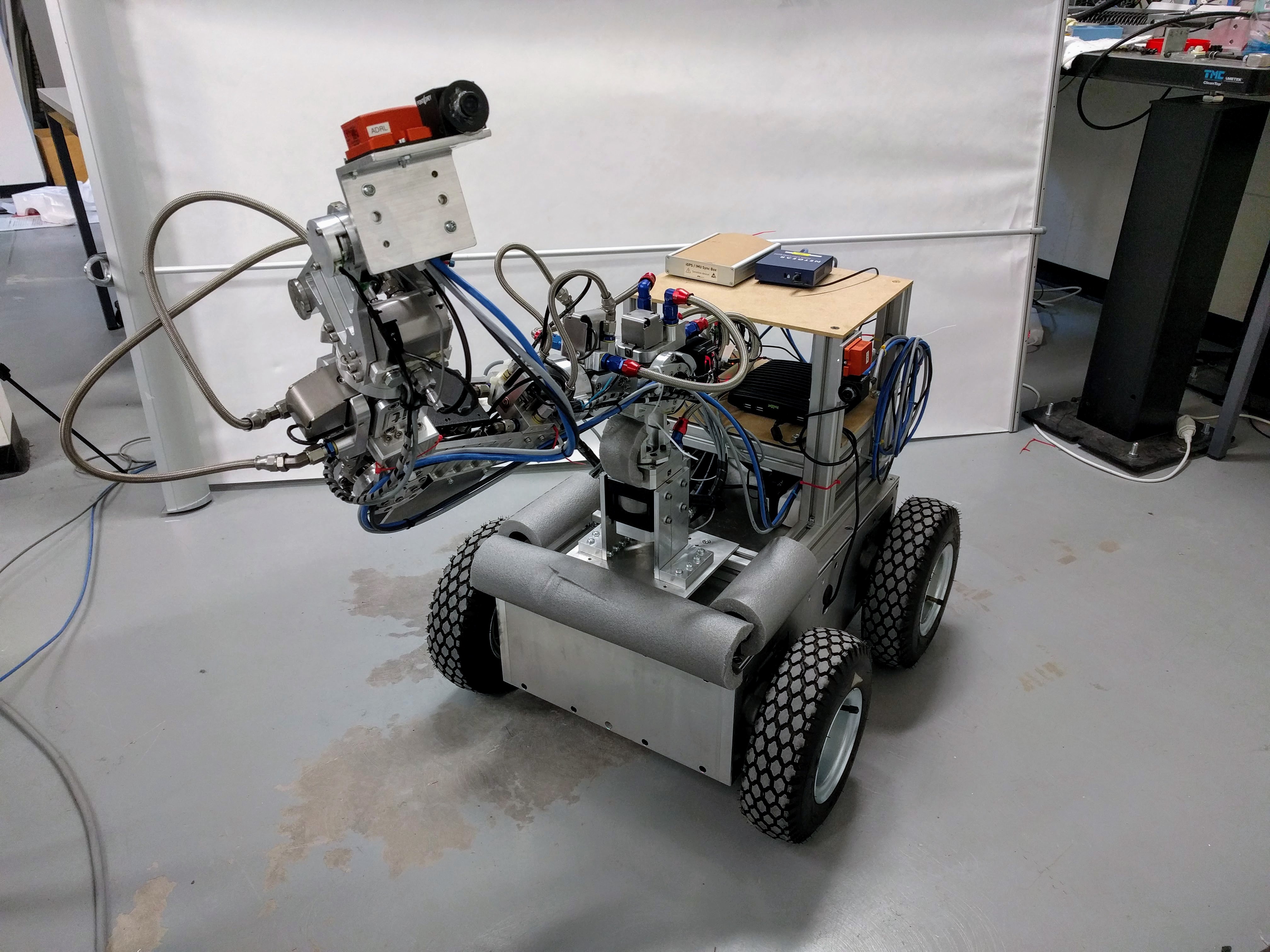}
	\caption{IFmini, a hydraulically actuated manipulator on a skid-steer base, equipped with a camera and IMU mounted both on the base and end-effector.}
	\label{fig:cf_ifmini}
\end{figure}

\subsection{Prior Work} \label{sec:cf_prior_work}
Most robotic systems today use filter-based sensor fusion algorithms for generating real-time state estimates for use in motion control. Extended Kalman filters (EKF) are popular for their very low computational overhead, low memory requirements, and ease of implementation~\cite{li2013high,lynen2013robust,bloesch2013state,xinjilefu2014decoupled}. In systems where non-linearities and non-uniform noise distributions are more dominant, other types of filters like the unscented Kalman filter (UKF) and particle filter are often used~\cite{bloesch2013slippery,fallon2014drift}. Filter-based sensor fusion has multiple well-known weaknesses, though. The sequential process-then-update nature of filtering schemes places a restriction on the structure of the sensing system. An explicit process model which models the evolution of the complete state from one time instance to the next is required~\cite{bloesch2018two}. Additionally, multiple process measurements which act on the same portion of the state cannot be directly incorporated into the filter. Finally, since all measurements are processed sequentially, with the information provided by each measurement summarized by its immediate linearization and update to the filter state, estimate errors can bias the influence of future measurements and future measurements cannot be used to retrospectively improve the accuracy of past estimates.

More complex sensor fusion schemes facilitate the improvement of estimates after future measurements have been received. Fixed-lag smoothers achieve this by iteratively taking forward and backward recursive passes over a batch of states which slides over time. Fixed-lag smoothers still enforce the restrictions in sensing system structure mentioned above, however. MHEs maintain an active batch of estimated states whose parameters are optimized simultaneously at each time-step. This allows the estimate of the state at a certain time to be repeatedly improved as the state marches back through the estimated batch of states. MHEs have been previously shown to outperform EKFs in monocular simultaneous localization and mapping (SLAM)~\cite{strasdat2010real} and in monitoring chemical processes~\cite{haseltine2005}. In a previous work, we showed that they can also provide smoother predictive estimates for use in high-frequency real-time robot controllers~\cite{sandy2017dynamically}. In~\cite{bloesch2018two}, the two-state implicit filter is proposed to relax the sensing system design constraints imposed by filter-based approaches. It is very similar to a MHE with a batch size of two, but updates to the estimated parameters are obtained using a recursive-style solver with worse performance in the presence of non-linearities than the solvers typically employed in MHEs. In~\cite{diehl2009efficient}, an open source MHE implementation is presented that was developed to maximize computational efficiency through automatic code generation and the use of specialized solvers. It was developed for applications with very limited computational resources and it is not clear how effective it would be for use in complex robotic systems. It also does not provide the full flexibility in sensing system design offered by MHEs because it runs an EKF at the front of the optimized batch of states to perform marginalization.


The biggest downside of using MHEs for online state estimation is their high computational cost relative to filter-based approaches. Over the last 10 years, multiple open-source packages for non-linear-least-squares optimization have been developed to support bundle adjustment and SLAM applications~\cite{kummerle2011g, ceres-solver, dellaert2012factor}. While these solvers were not developed with online usage in mind, their scalability and computational efficiency make them valuable tools in online applications as well. Okvis is a point-feature-based online visual-inertial odometry system that uses the Ceres Solver~\cite{ceres-solver} for non-linear-least-squares optimization in a MHE~\cite{leutenegger2015keyframe}. The sensor fusion algorithm employed is very similar to the one presented here, though we generalize it to support the fusion of an arbitrary number of general sensors and process models. iSAM2, also designed for SLAM applications, takes a different approach and estimates the full state history over time by efficiently factoring the probabilistic constraints between estimated parameters as they are incrementally induced by the measurements received~\cite{kaess2012isam2}. While this method works well in SLAM problems, where links between states and landmarks are relatively sparse over time, it is not clear how it would transfer to more general sensor fusion problems for robot state estimation, where links between states and other estimated parameters are often persistent.

\subsection{Contributions and Paper Structure} \label{sec:cf_contributions}
The main contributions of this work are as follows. We present a general framework for online sensor fusion that allows for the easy incorporation of additional sensors, multiple non-synchronized process measurements, and which can leverage additional computing resources to improve estimate quality. Our C++ implementation is made available open-source online\footnote{\url{https://bitbucket.org/tsandy/confusion}}. Additionally, we demonstrate a novel state estimator for mobile manipulation using dual visual-inertial sensors at the robot's base and end-effector, showing that the additional sensors improve localization accuracy versus ground truth measurements. 

This paper is structured as follows. The theory of the underlying sensor fusion problem used in ConFusion is presented in Section~\ref{sec:cf_mhe}. Our open-source implementation and its features are explained in Section~\ref{sec:cf_software}. Experimental results using ConFusion for visual-inertial tracking and whole body state estimation of a mobile manipulator are presented in Section~\ref{sec:cf_experiments}. Finally, conclusions and an outlook to future work are given in Section~\ref{sec:cf_conclusion}. 

\section{Moving-Horizon Estimation} \label{sec:cf_mhe}
We consider the problem of estimating the state of a robot at discrete instances of time ($\V{x}_t$) given a heterogeneous set of measurements. Measurements provide either information about the state of the robot at a specific time instance (such update measurements are written $\V{\hat{u}}_t$) or information relating the time evolution from one state instance to the next (a chain of such process measurements linking states $\V{x}_{t_i}$ and $\V{x}_{t_j}$ is written $\V{\hat{p}}_{t_i:t_j}$). In addition to the state of the robot, time-invariant (or static) parameters ($\V{s}$) are also estimated. These might be sensor intrinsic or extrinsic calibrations, or in the case of robot localization, the position of stationary references in the robot's environment. Update and process models of the following form are used to relate measurements received to the estimated parameters.
\begin{align*}
\V{u}_t &= g(\V{x}_t,\V{s}) \\
\V{x}_{t_j} &= h(\V{x}_{t_i},\V{\hat{p}}_{t_i:t_j},\V{s})
\end{align*}
The $g$ and $h$ functions typically only involve a subset of the state and static parameters and can be non-linear in the estimated parameters and measurements. The Markov property is implicitly assumed, as process chains can only link successive states.
Otherwise, no specific structure is assumed for these models, though the full observability of the estimated parameters is required from the full set of active measurements and models to achieve good estimator performance.

Considering a batch of $N$ states, determining the optimal set of parameters can be formulated as a least-squares problem of the following form.
\begin{align}
\begin{split}
\{ \V{x}_{t_0:t_N}^*, &\V{s}^* \} = \argmin_{\V{x}_{t_0:t_N}, \V{s}} \ \Bigg[ \ \Big\lVert 
\begin{matrix}
\check{\V{x}}_{t_0} \boxminus \V{x}_{t_0} \\ \check{\V{s}} \boxminus \V{s}
\end{matrix} \Big\rVert ^2_{W_M} + \\
&\sum_{i=0}^{N} \left( \sum_{\V{\hat{u}} \in \V{\mathcal{U}}_{t_i}}
\lVert \V{\hat{u}} \boxminus g(\V{x}_{t_i},\V{s}) \rVert ^2_{W_{\hat{u}}} \right) + \\
& \sum_{i=1}^{N} \left( \sum_{\V{\hat{p}} \in \V{\mathcal{P}}_{t_{i-1}:t_i}}
\lVert \V{x}_{t_i} \boxminus h(\V{x}_{t_{i-1}},\hat{\V{p}},\V{s}) \rVert ^2_{W_{\hat{p}}} \right) \Bigg]
\label{eq:cf_batch_problem}
\end{split}
\end{align}
The first term in the function being minimized captures any prior knowledge about the initial value of the first state and static parameters $(\check{\V{x}}_{t_0},\check{\V{s}})$. We use the notation $\lVert \V{a} \rVert^2_B = \V{a}^T B \V{a}$. When assuming that uncertainty in the initial state and the update measurement noise are normally distributed, the weighting matrices of the prior knowledge and update measurement residuals are chosen to be the inverse covariance of those quantities. Assuming normally distributed process noise, the weighting for the process chain residuals is the inverse covariance of the state resulting from forward propagating the estimate of the preceding state (starting with no uncertainty on the preceding state) through the process measurements. With these choices for the weighting matrices, this sum-of-squares cost function is therefore made up of unit-less terms, each representing a probabilistic quantity. We use $\boxminus$ operators here to indicate that the distance between any non-Euclidean quantities (e.g. for parameters representing rotations in 3d space) is computed in the tangent space centered at the current value of those parameters.

Equation~(\ref{eq:cf_batch_problem}) is solved iteratively using nonlinear least squares optimization. At each iteration, all of the residuals are linearized about the current value of the estimated parameters. If we define $\V{x}$ to be the a stacked vector of all of the estimated state and static parameters, $\V{e_i} = W^{\frac{1}{2}} f_i(\V{x})$ such that $\V{e_i}^T \V{e_i} = \lVert f_i(\V{x}) \rVert^2_W$, $\V{e} = [\V{e_0}^T, ... , \V{e_m}^T]^T$, and $\V{J} = \frac{\delta \V{e}}{\delta \V{x}}$, the right side of (\ref{eq:cf_batch_problem}) can be approximated as a linear least squares problem,
\begin{equation}
\argmin_{\delta \V{x}} \lVert \V{J} \delta \V{x} + \V{e} \rVert^2,
\label{eq:cf_linear_problem}
\end{equation}
where we now consider solving for the optimal increment to apply to the estimated parameters, e.g. $\V{x}'=\V{x} \boxplus \delta \V{x}$, where $\boxplus$ is used to generalize addition to the case of incrementing over-parameterized quantities by a $\delta \V{x}$ expressed in the tangent space of the current estimate. By defining $\V{H}=\V{J}^T\V{J}$ and $\V{b}=-\V{J}^T \V{e}$, solving (\ref{eq:cf_linear_problem}) is equivalent to solving the so-called \emph{normal equations} of the problem, $\V{H} \delta \V{x} = \V{b}$. Since $\V{H}$ is inherently sparse due to the assumed Markov property, this equation is solved at each iteration using sparse Cholesky factorization. The Levenberg-Marquart algorithm is used to provide more robustness in optimization than the Gauss-Newton algorithm, while achieving convergence in fewer iterations than gradient descent~\cite{barfoot2017state}. Evaluating the individual residuals is done in parallel over a user-specified number of threads, though solving the normal equations at each iteration is done from a single thread. 

In the presence of outlier measurements or non-normal noise distributions, the squared residual terms can be enclosed in a loss function (e.g. Huber loss) to decrease their influence in the presence of large residual values. This results in a robustified iteratively re-weighted least squares optimization problem.

\subsection{Marginalization}
\begin{figure}
	\centering
	\includegraphics[width=0.8\columnwidth]{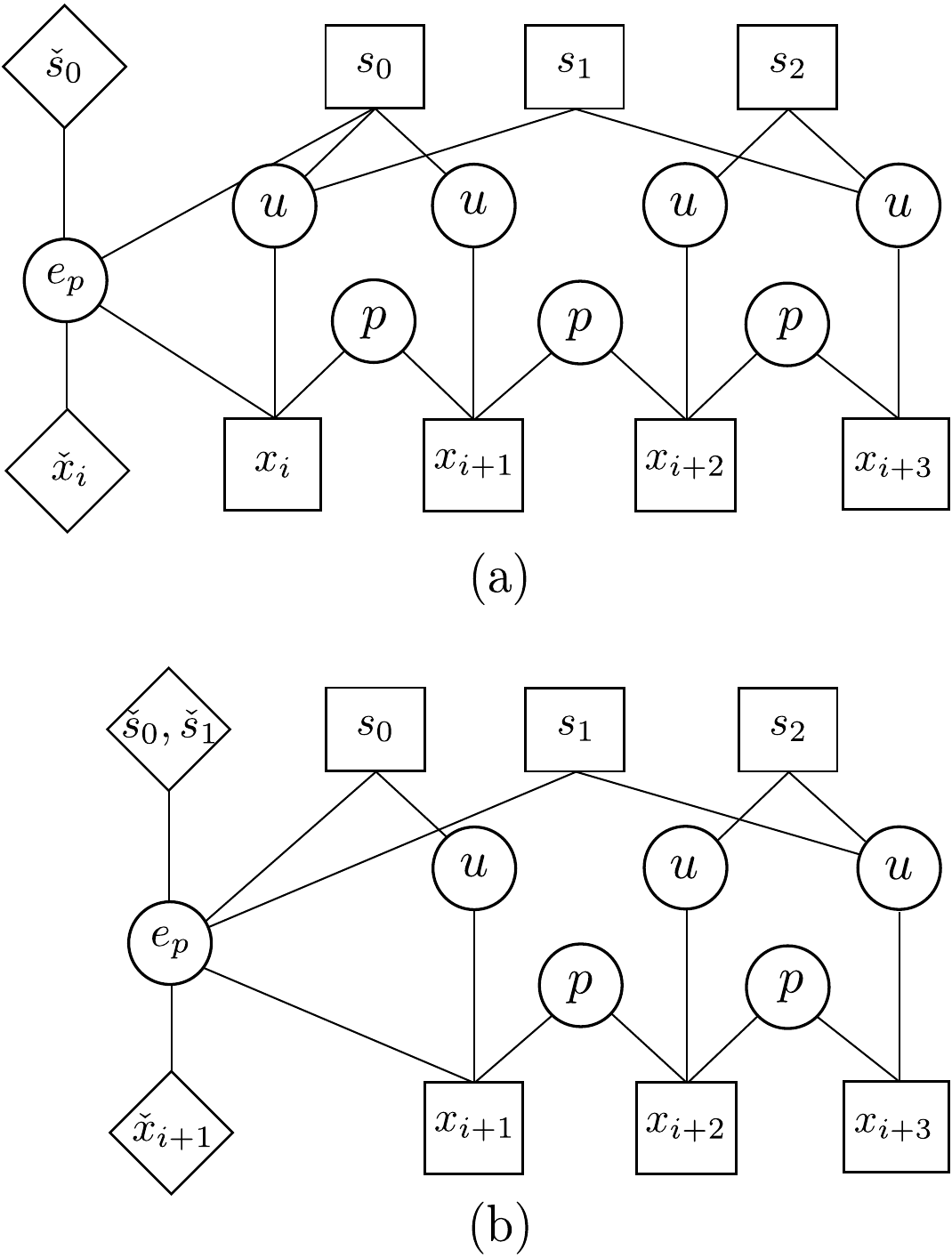}
	\caption{Diagrams showing the structure of a MHE problem before and after marginalization. Estimated parameters are drawn as squares, residuals as circles, and marginalized parameters as diamonds. Drawn edges represent the dependence of residuals on the connected parameters.}
	\label{fig:cf_mhe_diagram}
\end{figure}
It is not feasible to solve (\ref{eq:cf_batch_problem}) continuously online since the number of estimated parameters grows linearly over time. To deal with this, measurements are removed (or marginalized) out of the problem once their influence on the estimated parameters becomes sufficiently small. The information contained in the marginalized measurements is approximated in what we call the prior constraint. This residual function relates the remaining estimated parameters which were linked to the marginalized measurements to their values at the time of marginalization. The problem structure before and after marginalizing out the measurements connected to a state is shown graphically in Fig.~\ref{fig:cf_mhe_diagram}. The MHE optimization problem considering a receding horizon (or batch) of $N$ states is
\begin{align}
\begin{split}
\{ \V{x}_{t_i:t_{i+N}}^*&, \V{s}^* \} = \argmin_{\V{x}_{t_i:t_{i+N}}, \V{s}} \  
\lVert \V{e}_{p}(\check{\V{x}}_{t_i},\V{x}_{t_i},\check{\V{s}},\V{s}) \rVert^2 + \\
&\sum_{j=i}^{i+N} \left( \sum_{\V{\hat{u}} \in \V{\mathcal{U}}_{t_j}}
\lVert \V{\hat{u}} \boxminus g(\V{x}_{t_j},\V{s}) \rVert ^2_{W_{\hat{u}}} \right) + \\
&\sum_{j=i}^{i+N} \left( \sum_{\V{\hat{p}} \in \V{\mathcal{P}}_{t_{j-1}:t_j}}
\lVert \V{x}_{t_j} \boxminus h(\V{x}_{t_{j-1}},\hat{\V{p}},\V{s}) \rVert ^2_{W_{\hat{p}}} \right).
\label{eq:cf_mhe_problem}
\end{split}
\end{align}

Estimated parameters are marginalized out of the problem using block Gaussian elimination on the normal equations of the underlying least squares problem. To marginalize out some subset of the estimated parameters, $\V{x}_{m}$, we first identify the remaining estimated parameters ($\V{x}_{l}$) which are directly linked to the marginalized parameters by a  residual. A sub-problem is then built, made up of all the residuals in which $\V{x}_{m}$ appears. The normal equations of this subproblem can be reordered as follows.
%
%
\begin{equation*}
\left[
\begin{array}{cc}
\V{H}_{mm} & \V{H}_{ml} \\
\V{H}_{lm} & \V{H}_{ll}
\end{array}
\right] \left[
\begin{array}{c}
\delta\V{x}_{m} \\
\delta\V{x}_{l}
\end{array}
\right] = \left[
\begin{array}{c}
\V{b}_{m} \\
\V{b}_{l}
\end{array}
\right]
\end{equation*}
The subscripts of $\V{H}$ and $\V{b}$ reflect the row and column-wise associations to $\V{x}_{m}$ and $\V{x}_{l}$. By left-multiplying both sides of the system of equations with the Schur complement of the partitioned $\V{H}$ matrix, we obtain
%
%
\begin{equation*}
\left[
\begin{array}{cc}
\V{H}_{mm} & \V{H}_{ml} \\
\V{0} & \V{H}^*
\end{array}
\right] \left[
\begin{array}{c}
\delta\V{x}_{m} \\
\delta\V{x}_{l}
\end{array}
\right] = \left[
\begin{array}{c}
\V{b}_{m} \\
\V{b}^*
\end{array}
\right]
\end{equation*}
where
\begin{align*}
\V{H}^*&=\V{H}_{ll}-\V{H}_{lm}\V{H}_{mm}^{-1}\V{H}_{ml} \\
\V{b}^*&=\V{b}_{l}-\V{H}_{lm}\V{H}_{mm}^{-1}\V{b}_{m}
\end{align*}
and $\delta\V{x}_{l}$ can be found independent of $\delta\V{x}_{m}$.

The resulting prior constraint is formulated as
\begin{equation*}
\V{e}_p = \V{J}_p (\check{\V{x}}_{l} - \V{x}_{l}) - (\V{J}_p^T)^\dagger \V{b}^*,
\end{equation*}
where $\check{\V{x}}_{l}$ is the value of $\V{x}_{l}$ at the time of marginalization. $\V{J}_p$ is obtained from $\V{H}^* = \V{J}_p^T\V{J}_p$ via LU decomposition and $(\V{J}_p^T)^\dagger$ is computed using the Moore-penrose pseudo-inverse. When taking the Schur complement, $\V{H}_{mm}$ is inverted using Cholesky decomposition. 
%

In the case that new static parameters are being continuously added to the problem over time, e.g. while performing SLAM in unknown environments, the prior constraint will also grow over time. To control the size of the prior constraint, static parameters can be factored out of it as desired. This is similarly done by applying the Schur complement to the $\V{H}^*$ and $\V{b}^*$ computed in the previous marginalization step, but reordered such that the parameters to be removed appear in the left-most columns. As was done in our previous work in object-based SLAM~\cite{sandy2018object}, the parameters of the first state which are linked to the prior constraint can be similarly factored out of the prior constraint to allow sensor fusion to be stopped and re-started while maintaining the accumulated relative certainties in the static parameters in a probabilistically consistent way.

\subsection{Problem Structure}
The proposed estimator generates maximum a posteriori estimates of the following form~\cite{barfoot2017state}:
\begin{equation*}
	\{\V{x}_{i:i+N}^*, \V{s}^*\} = \argmax_{\V{x}_{i:i+N}, \V{s}} \rho \left( \V{x}_{i:i+N}, \V{s} | \V{\mathcal{U}}_{i:i+N}, \V{\mathcal{P}}_{i:i+N}, \check{\V{x}_{i}}, \check{\V{s}} \right),
\end{equation*}
where $\V{\mathcal{U}}_{i:i+N}$ and $\V{\mathcal{P}}_{i:i+N}$ signify the set of update and process measurements received between times $t_i$ and $t_{i+N}$, and $\check{\V{x}_{i}}$ and $\check{\V{s}}$ are the values of those parameters from the time of the last marginalization. Although it is not explicitly computed when solving the MHE problem, the covariance of the generated estimates can be obtained by computing the inverse of $\V{H}$.

\begin{figure}
	\begin{center}
		\begin{tabular}{ r c c c c c c c c }
			& & $\V{c}$ & $\V{m}$ & $\V{x}_i$ & $\V{x}_{i+1}$  & $\V{x}_{i+2}$ & $\V{x}_{i+3}$ \\[0.2cm]
			$\V{c}$ & \ldelim[{6}{2mm}[] & X &  & \cellcolor{blue!20}X & \cellcolor{yellow!20}X & \cellcolor{yellow!20}X & \cellcolor{yellow!20}X & \rdelim]{6}{2mm}[] \\
			$\V{m}$ &  & & X & \cellcolor{blue!20}X & \cellcolor{green!20}X & \cellcolor{green!20}X & \cellcolor{green!20}X & \\
			$\V{x}_i$ & & \cellcolor{blue!20}X & \cellcolor{blue!20}X & \cellcolor{blue!20}X & X &  &  & \\
			$\V{x}_{i+1}$ & & \cellcolor{yellow!20}X & \cellcolor{green!20}X & X & X & X & & \\
			$\V{x}_{i+2}$ & & \cellcolor{yellow!20}X & \cellcolor{green!20}X & & X & X & X & \\
			$\V{x}_{i+3}$ & & \cellcolor{yellow!20}X & \cellcolor{green!20}X & & & X & X & \\
		\end{tabular}
	\end{center}
	\caption{Illustration of the sparsity structure of the $\V{H}$ matrix during operation, considering a batch of four states and static parameters made up of the robot calibration ($\V{c}$) and the map used for localization ($\V{m}$).}
	\label{fig:sparsity} 
\end{figure}
Fig.~\ref{fig:sparsity} shows the sparsity structure of $\V{H}$ during operation. For illustration, we divide the static parameters into those which make up the robot's map used for localization, $\V{m}$, which might be in the least the gravity alignment to a tracked external reference frame or at most a full SLAM map, and those that relate to the robot's internal calibration, $\V{c}$, which could be internal sensor offsets and biases. The lower right portion has a regular block diagonal structure induced by the Markov property between states. When calibrations are being optimized online, they will almost always generate persistent links to all states in the portion shown in yellow. The sparsity of the portion shown in green is dependent on the representation of the map used. There are typically persistent links between the map and states over time, but as the dimensionality of the map parameters grows, connections to specific map parameters become more sparse. The proposed sensor fusion scheme will create fill-in between the static parameters and the first state in the batch (the portion shown in blue) due to the marginalization of measurements connected to past states. As a result, the top-left corner of $\V{H}$ up to and including this blue section will have the same structure and sparsity as the inverse innovation matrix in a Kalman filtering setting, with the additional states in the batch contributing a more sparse ``tail". In Sec.~\ref{sec:cf_vi_tracking} we show how the use of a sparse linear solver leverages this sparsity to achieve reasonable computation scaling with the batch size. While alternative marginalization schemes have been proposed for large SLAM problems (e.g.~\cite{kaess2012isam2}), such methods are not applicable here because they would require significant sparsity in the yellow and green portions of $\V{H}$, which is often not the case for robot state estimation problems.

\section{Software Structure} \label{sec:cf_software}
Our C++ implementation of the proposed MHE sensor fusion framework, called ConFusion, is made available open-source online\footnote{\url{https://bitbucket.org/tsandy/confusion}}. ConFusion essentially wraps around the Ceres Solver~\cite{ceres-solver}, a well-established non-linear least squares solver, allowing users to take advantage of its support for auto-differentiation, robust loss functions, and a wide range of non-linear solvers. ConFusion provides a higher level API specialized for the operation of batch-based online state estimation for robots with arbitrary structure and combinations of sensors. Some implementation features are as follows:
\begin{itemize}
	\item ConFusion exposes functions to build and solve state estimation problems, and marginalize out states and static parameters as desired, without the user having to manipulate the internal MHE problem structure. ConFusion therefore requires no additional input from the user than would be required in a filter-based setting.
	\item ConFusion takes care of spawning new states and assigning measurements to states as they arrive, taking the burden of buffering and assigning measurements off of the user.
	\item We provide a base set of measurement models, to allow a user to quickly plug together a state estimator for their robot. We encourage users to contribute new measurement models to build up a library of models over time.
	\item Both static and state parameters can be dynamically switched to be optimized or constant at any time, giving full flexibility of when parameters are estimated online.
	\item Batch calibration (or bundle adjustment) problems can be built and solved using the same state and measurement models used for online sensor fusion. This can be used to periodically run offline calibration runs on longer trajectories where the observability of system calibration parameters is ensured.
	\item Utilities are provided to visualize the structure of the underlying MHE problem for debugging, and for logging and plotting data in Matlab for tuning and analysis.
\end{itemize}

\section{Experiments} \label{sec:cf_experiments}
We demonstrate the performance of ConFusion through two sets of experiments. First, we compare the performance of ConFusion to that of an iterated extended Kalman filter (IEKF) on the visual-inertial tracking of a sensor-head and show the impact of batch size on the resulting real-time estimate accuracy and smoothness. Second, we show the extendability of ConFusion by performing whole-body sensor fusion on a mobile manipulator using different combinations of sensors. Our visual-inertial sensor heads are comprised of an Xsens MTi-100 inertial measurement unit (IMU), delivering accelerometer and gyroscope measurements at 400 Hz, and a PointGrey Blackfly monochrome camera providing images at 10 Hz and 5 mega-pixel resolution. As visual references, we use stationary AprilTag~\cite{olson2011tags} fiducial markers. Experiments were run using a standard laptop~(Intel i7-4800MQ).

In both data-sets, we compare the estimated trajectories to ground truth measurements captured with a Leica Tracker (AT960), which provides full pose measurements of a tracked frame at 50 Hz and with a nominal accuracy of less than 0.1~mm and \SI{3.5e-4}~rad. A Leica T-Mac (TMC30-F) was mounted rigidly to our visual-inertial sensor unit. This sensor unit was then either moved by hand (Section~\ref{sec:cf_vi_tracking}) or mounted on the end-effector of our robot (Section~\ref{sec:cf_mobile_manip}). The Leica Tracker was placed about 6~meters from the experimental workspace.
%
\subsection{Visual-Inertial Tracking} \label{sec:cf_vi_tracking}
In this section, we show the performance of ConFusion in visual-inertial tracking, for which there is a large body of prior work. Many works have investigated using extended Kalman filters for fiducial marker map building~\cite{neunert2016open}, visual odometry~\cite{mourikis2007multi,forster2017manifold}, and object tracking~\cite{sandy2018object}. Other works have considered using batch-based methods for visual-inertial odometry~\cite{leutenegger2015keyframe} and SLAM~\cite{mur2017visual,qin2018vins}. While the majority of these works pursue the goal of achieving bounded accuracy over large environments and long trajectories, our work here considers the goal of generating high-accuracy and smooth state estimates within small environments. This scenario is more consistent with the requirements for articulated mobile robots interacting with their environment. The reader is referred to our previous work~\cite{sandy2017dynamically} for the sensor models and conventions employed here.

\begin{figure}
	\centering
	\includegraphics[trim=20 0 30 0, clip,width=\columnwidth]{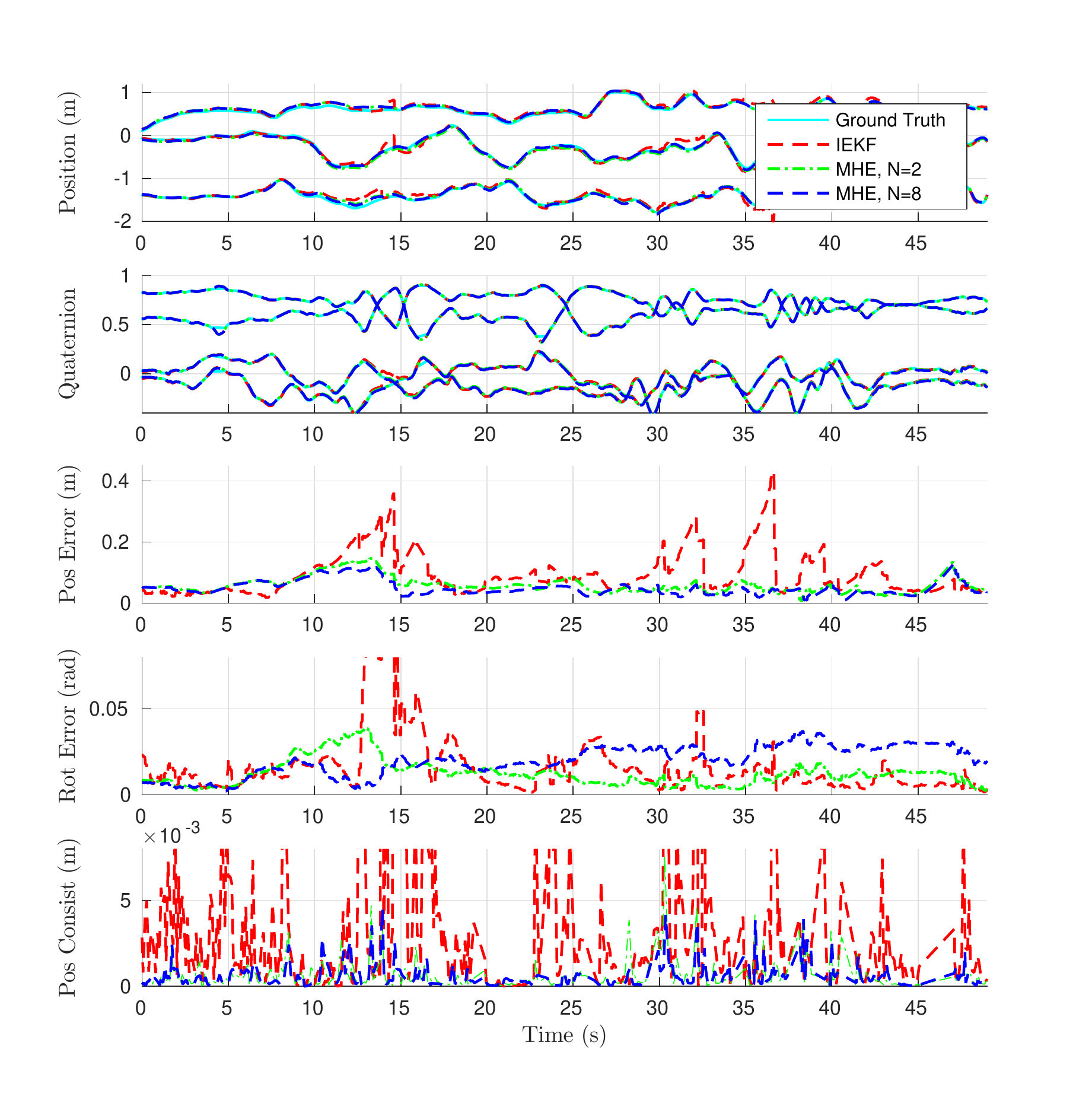}
	\caption{Accuracy of the generated real-time sensor-head trajectories using an IEKF, ConFusion with a batch size of two run on a single thread, and ConFusion with a batch size of 8 run on two threads. The MHE-generated estimates are significantly more accurate and smoother than the IEKF estimates. Peaks in the IEKF plots in the bottom two plots are cut-off for clarity in the other plots.}
	\label{fig:cf_vi_gterror}
\end{figure}

Fig.~\ref{fig:cf_vi_gterror} shows the accuracy of the estimated IMU pose in the presence of aggressive motions using different sensor fusion schemes. The estimates consider an online mapping scenario with no prior knowledge of the fiducial marker poses. To simulate the generation of real-time estimates for use in robot motion control, the estimates were generated by playing back the recorded data in real-time, generating the estimates online, and then forward-propagating them through the more recent IMU measurements up to the time of the most recent IMU measurement. The frame offsets which fit the generated trajectories to the ground truth data were calibrated in a batch problem using all of the measurements in the dataset. It can be seen that the MHE estimates generated by confusion are both more accurate and smoother than those of the IEKF. Although the rotational error is higher when using a bigger batch size, the fact that the rotational error of the trajectory with N=8 in Fig.~\ref{fig:cf_vi_gterror} is a slowly increasing offset leads us to believe this is caused by a shift in the estimator's global frame during operation, causing a misalignment with the calibrated offsets to the ground truth measurement system. 
The bottom subplot shows the consistency of the real-time forward-propagated position estimates. This is computed as the difference between the delayed estimate from ConFusion and forward-propagated estimate from the same time. This shows how well the IMU process model and estimated intrinsic parameters fit the true evolution of the sensor-head motion. Once again, you see that the ConFusion estimates outperform the IEKF. These estimated trajectories can be seen in the accompanying video\footnote{\url{https://youtu.be/J6up9Eq9Sxc}}.

\begin{figure}
	\centering
	\includegraphics[trim=0 0 0 0, clip,width=\columnwidth]{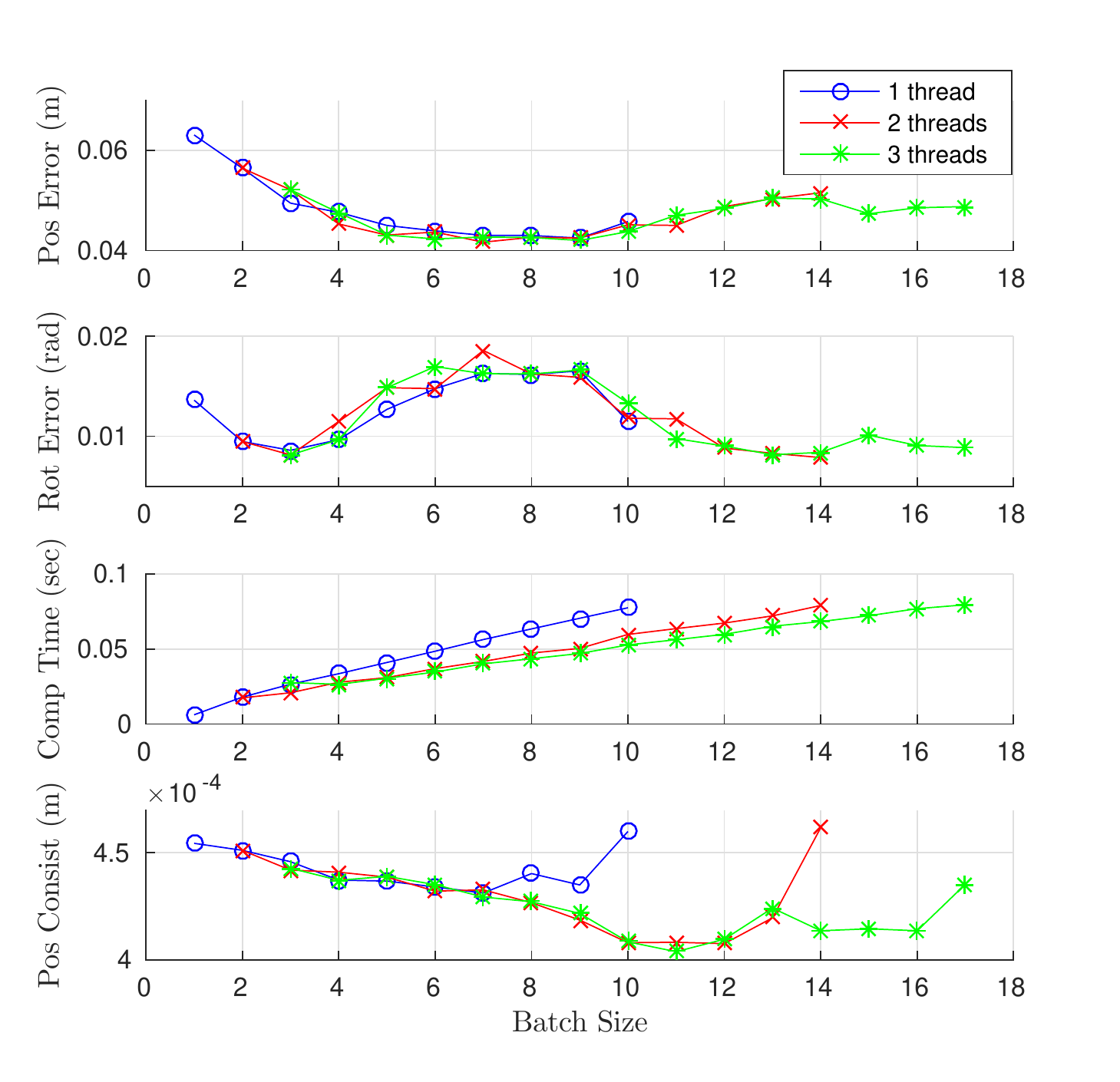}
	\caption{The impact of batch size on the visual-inertial tracking performance. The RMS error in the real-time estimates, versus ground truth, is shown in the top two subplots. The sensor fusion computation time is shown in the third subplot. The fourth subplot shows the RMS translational error between the delayed optimized state estimates and the real-time estimates at those times, showing the consistency of the forward propagation through the IMU process model.}
	\label{fig:cf_vivsbatchsize}
\end{figure}

Fig.~\ref{fig:cf_vivsbatchsize} shows how the estimator performance is impacted by the batch size and number of cores used for computation in ConFusion. You can see that the tracking accuracy increases with batch size up to some point, and then slightly decreases. This shows the trade-off between using a larger batch to decrease the estimate errors incorporated into the prior constraint over time, and the increased computation time and latency in the generated estimates. We believe that the rise in rotational error with batch size is once-again due to the misalignment with the calibrated offset to the ground truth system and does not reflect a true degradation in accuracy. The third subplot shows the average computation time for building and solving the MHE problems and performing marginalization. With each number of threads, the batch size was increased until the computation time started to consistently overrun the image frame rate of 10 Hz. You can see that increasing the batch size has a nearly linear impact on the computation time, showing that the sparsity of the system Jacobian is effectively being leveraged to improve computational efficiency. The use of additional cores allows for the usage of a bigger batch of states, though the impact decreases with larger batch sizes since the solver step of the algorithm only uses a single thread. Finally, the bottom subplot shows the root-mean-square~(RMS) value of the same consistency measure introduced in the previous paragraph. The use of a larger batch does indeed have a smoothing effect, improving the consistency measure by more than 10 percent. This is important for use in real-time control systems as any inconsistency in the generated real-time estimates is seen as a disturbance by the robot's control system.


\subsection{Mobile Manipulator Whole-Body State Estimation} \label{sec:cf_mobile_manip}
We next consider the problem of estimating the state of a mobile manipulator with ConFusion. We use a new robot, called IFmini (Fig.~\ref{fig:cf_ifmini}), which consists of a 6 degree of freedom hydraulically actuated manipulator, designed and built by the Italian Institute of Technology~\cite{rehman2015}, mounted on top of a four wheeled differential drive base, called the Supermegabot by Inspectorbots. IFmini is a more dynamic, albeit smaller, successor to the In situ Fabricator (IF), designed for performing building construction tasks directly on the construction site~\cite{giftthaler2017mobile}. 

In the interest of performing dynamic manipulation tasks with high accuracy, we would like to estimate the pose and velocity of IFmini's end-effector within its environment. For use in control, these estimates should be available at high rate and with minimal latency. To support whole-body model-based control, it is desirable to additionally estimate the pose and velocity of the base. Here we show that we can generate accurate real-time estimates of the base and end-effector states using ConFusion. Building on our past results~\cite{sandy2017dynamically}, the consistency of our MHE estimates allows for the accurate forward propagation of our state estimates through the IMU measurements up to real time for use in control.

In order to support whole-body sensor fusion, and also to investigate which sensors are most valuable for generating high-accuracy real-time estimates, we have equipped IFmini with a camera and IMU on both the base and end-effector. In this configuration, the robot state is simply two times the state considered for visual-inertial tracking in the previous section, with both cameras localizing the robot within the same map. We additionally fuse joint angle measurements and a model of the arm's kinematics to relate the relative poses of the base and end-effector as an additional update measurement. Finally, we consider a simple differential drive kinematic model of the base and wheel speed measurements arriving at 75~Hz as an additional process model on the time evolution of the estimated base pose. The modeled confidence in the base motion model is significantly lower when the wheels are turning than when they are stationary since our skid steer base experiences significant wheel slip while driving. This motion model therefore mainly helps enforce that the base remains stationary when the wheels are not turning.

The relative pose of the arm's base frame with respect to the base IMU, the relative pose of the arm's end-effector with respect to the end-effector IMU, and the joint angle biases for the middle four arm joints\footnote{Biases for the joints closest to the base and end-effector are not considered because they are compensated for in the relative poses of the IMUs with respect to the manipulator.} are added to the estimation problem as static parameters to align the sensors with the arm kinematic model. Although ConFusion supports solving for these parameters online, we find that it is most effective to calibrate them separately in a batch problem and leave them fixed during online operation. This ensures that their values do not over-fit configuration-specific errors within the sensing system (e.g. camera intrinsic calibration inaccuracy) while the robot is stationary. Online calibration can nevertheless be useful in cases where, for example, sensors must be removed and re-mounted on a robot often or relative encoders are used for joint odometry. In these cases, small changes in the sensor calibration can be quickly identified without requiring that an open-loop calibration routine is run before the robot's state estimator is turned on.

\begin{figure}
	\centering
	\includegraphics[trim=0 0 0 0, clip,width=\columnwidth]{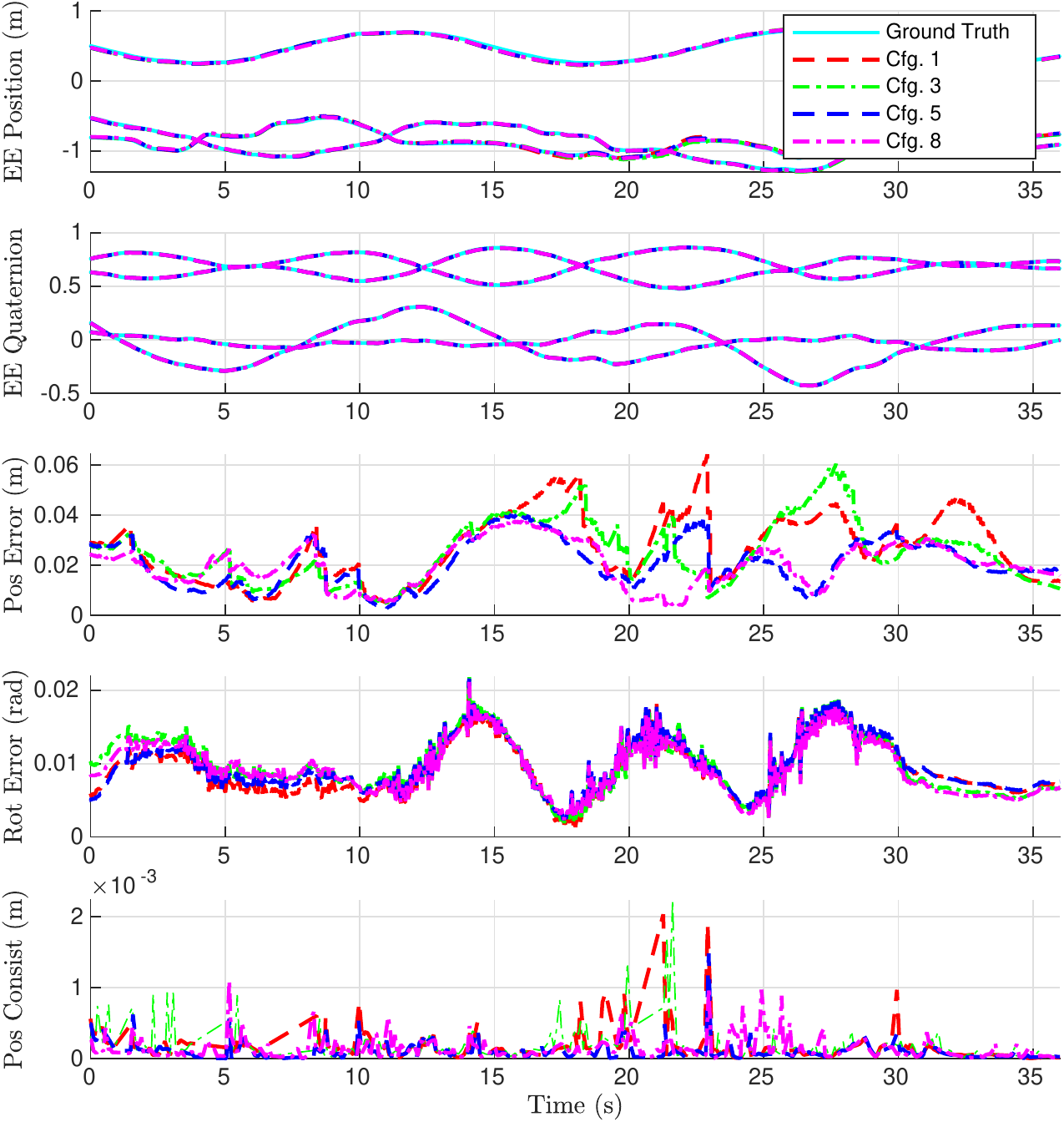}
	\caption{IFmini end-effector pose estimate accuracy versus ground truth using different sets of sensors in the state estimator. The sensors used in each configuration are shown in Table~\ref{tab:cf_ifmini_table}.}
	\label{fig:cf_ifmini_gterror}
\end{figure}

Estimated robot trajectories were generated from the same dataset using measurements from different combinations of sensors. While whole-body estimates were generated, only the end-effector portion of the state can be compared to the ground truth measurements from the Leica Tracker. This is a good measure of the state estimate quality for manipulation tasks, however, since they require controlling a robot's end-effector motion relative to its environment. Table~\ref{tab:cf_ifmini_table} shows the RMS end-effector estimated pose error and the sensors used for the different runs. The consistency of the estimated end-effector position is also shown, similar to in Fig.~\ref{fig:cf_vivsbatchsize}. Fig.~\ref{fig:cf_ifmini_gterror} shows four selected trajectories in more detail to get a sense of the robot motion during the experiment. All estimates were generated using a batch size of 5 running on two threads. Once again, estimates were forward propagated up to the time of the most recent IMU measurement to simulate real-time operation on the robot. The sensor extrinsics, offsets to the ground truth measurement frames, and a map of the fiducial marker poses were first calibrated in a batch problem on a different dataset and then held constant for this experiment. The performance of the state estimator running on the robot in real-time and being used to close an end-effector task-space motion control loop, is shown in the accompanying video\footnote{\url{https://youtu.be/J6up9Eq9Sxc}}.

The flexibility of ConFusion allows these different sensor configurations to be used with the same state representation and by simply changing an enumeration specifying which sensors are active. Not all of the sensor setups investigated could be used in a filter-based framework as, in configuration 1, the base portion of the state was not observed, in configuration 3, the base states have no linked process measurements but are only linked to the rest of the problem through the update measurements and static parameters, and, in configuration 6, the base portion of the state is constrained by two asynchronous process models.

We observe that in most cases adding additional sensors improves the state estimator performance. With visual-inertial sensing on the end-effector alone, there are portions of the trajectory where no fiducials are visible (e.g. 20-22 sec), resulting in large errors due to uncorrected drift. Adding the camera mounted on the robot base provides an additional viewpoint for viewing the fiducials, making it more likely that visual references are always visible. The use of the base camera and arm odometry also makes the extrinsic calibration of the cameras with respect to the manipulator and the joint angle biases immediately observable from the observation of the same reference in both cameras. Using the base IMU causes a degradation of performance in most cases. We observe that this is likely because the base IMU extrinsic calibration and sensor biases are not well observed during the predominantly planar base motion in the recorded data files. The system calibration gives different values for this extrinsic calibration when run on different datasets. Although this suggests that it is more effective to put an IMU on the end-effector than on the base for operation on flat ground, we plan to investigate this further in future work. 
\begin{table*}
	\begin{center}
		\begin{tabular}{ | c | c c c c c c | c c c | }
			\hline
			\multirow{2}{0.6cm}{\centering Cfg} &
			\multirow{2}{0.8cm}{\centering EE Cam} &
			\multirow{2}{0.8cm}{\centering EE IMU} &
			\multirow{2}{0.8cm}{\centering Base Cam} &
			\multirow{2}{0.8cm}{\centering Base IMU} &
			\multirow{2}{1cm}{\centering Arm Odometry} &
			\multirow{2}{2cm}{\centering Base Motion Model} & 
			\multirow{2}{0.8cm}{\centering Pos [m]} &
			\multirow{2}{0.8cm}{\centering Rot [rad]} &
			\multirow{2}{1.5cm}{\centering Pos Consist [m]} \\
			& & & & & & & & & \\[0.05cm] \hline 
			1 & X & X & & & & & 0.0298 & 0.0131 & 0.000280 \\
			2 & X & X & & & X & X & 0.0295 & 0.0129 & 0.000266 \\
			3 & X & X & & X & X & & 0.0284 & 0.0144 & 0.000378 \\
			4 & X & & X & X & X & & 0.0434 & 0.0324 & 0.000139 \\
			5 & X & X & X & & X & & \textbf{0.0201} & \textbf{0.0095} & \textbf{0.000137} \\
			6 & X & X & X & X & X & & 0.0204 & 0.0103 & 0.000248 \\
			7 & X & X & X & & X & X & \textbf{0.0201} & \textbf{0.0095} & \textbf{0.000137} \\
			8 & X & X & X & X & X & X & 0.0203 & 0.0103 & 0.000251 \\  \hline			
		\end{tabular}
	\end{center}
	\caption{Root-mean-squared position and rotation errors and the consistency of the real-time position estimates (see Fig.~\ref{fig:cf_vivsbatchsize} caption) during whole-body motion of IFmini, versus sensor configuration.}
	\label{tab:cf_ifmini_table} 
\end{table*}

\addtolength{\textheight}{-4.1cm}

\section{Conclusion} \label{sec:cf_conclusion}

In this paper, we have presented a general framework for batch-based online robot sensor fusion, released as an open-source software package called ConFusion. We have demonstrated its ability to generate more accurate estimates than filtering-based approaches, leverage additional computing resources to improve estimator performance, and support complex sensor setups on articulated mobile robots. As future work, we plan to investigate ways to parallelize the linear solver step in the MHE optimization problem and more principled strategies for determining when measurements and parameters should be marginalized out of the problem. With regard to the IFmini system, we plan to use the demonstrated state estimator to guide the execution of dynamic high-accuracy manipulation tasks and to investigate if the articulated dual-camera setup can be used to improve mapping and localization accuracy for natural-feature-based SLAM.

\section*{Acknowledgement}

The authors would like to give special thanks to Prof. Andreas Wieser and Robert Presl for providing and operating the Leica Tracker for acquiring ground truth measurements.


\bibliographystyle{ieeetr}
\bibliography{refs}

\end{document}